# Automated Detecting and Placing Road Objects from Street-level Images

Chaoquan Zhang[a,b], Hongchao Fan[a]*, Wanzhi Li[b], Bo Mao[c] and Xuan Ding[b]

[a]*Department of Civil and Environmental Engineering, Norwegian University of Science and Technology, Trondheim, Norway;* [b]*School of Remote Sensing and Information Engineering, Wuhan University, Wuhan, China;* [c]*College of Information Engineering, Nanjing University of Finance & Economics, Nanjing, China*

*Correspondence: hongchao.fan@ntnu.no

# Automated Detecting and Placing Road Objects from Street-level Images

Abstract: Navigation services utilized by autonomous vehicles or ordinary users require the availability of detailed information about road-related objects and their geolocations, especially at road intersections. However, these road intersections are mainly represented as point elements without detailed information, or are even not available in current versions of crowdsourced mapping databases including OpenStreetMap (OSM). This study develops an approach to automatically detect road objects and place them to right location from street-level images. Our processing pipeline relies on two convolutional neural networks: the first segments the images, while the second detects and classifies the specific objects. Moreover, to locate the detected objects, we establish an attributed topological binary tree (ATBT) based on urban grammar for each image in an image track to depict the coherent relations of topologies, attributes and semantics of the road objects. Then the ATBT is further matched with map features on OSM to determine the right placed location. The proposed method has been applied to a case study in Berlin, Germany. We validate the effectiveness of our method on two object classes: traffic signs and traffic lights. Experimental results demonstrate that the proposed approach provides near-precise localization results in terms of completeness and positional accuracy. Among many potential applications, the output may be combined with other sources of data to guide autonomous vehicles.

## 1. Introduction

The rapid development of advanced driver assistance systems and autonomous vehicles in recent years has attracted the ever-growing interest in smart traffic applications. Such intelligent applications can provide detailed road asset inventories of all stationary objects, such as street furniture (traffic lights and signs, various poles, bench, etc.), road information (lanes, edges, shoulders, etc.), small façade elements (antennas, cameras, etc.), and other minor landmarks. However, these detailed road map productions are mainly generated by mobile mapping system (MMS), which requires high cost both in the investment of equipment and in labor-intensive post data processing. In addition, the data updating is again a huge challenge. For instance, official road maps suffer from a long update cycle that can last several months or even years (Kuntzsch et al. 2016).

Recent one decade has witnessed an explosion of geospatial data. An increasing number of crowdsourced geospatial data repositories/services allow volunteers to utilize information from various data sources when contributing data to a crowd-sourced platform. That is known as Volunteered Geographic Information (VGI) (Goodchild, 2007). Amongst them, OSM and Mapillary are the typical representatives of maps and street-level crowdsourcing platforms, respectively. The large amount of detailed map data provided by OSM not only enriches the data sources of map producing, but also supports and promotes data-driven (Hachmann et al. 2018; Melnikov et al. 2016) and data-intensive (Chang et al. 2016; Gao et al. 2017) spatial analysis. Additionally, literature (Neis et al. 2012) has shown that OSM road data in Germany and Netherlands can be comparable to official data. With the introduction of Mapillary in 2014, it has become the biggest and the most active crowdsourced street-level imagery platform around the world. Tens of billions of street view images covering millions of kilometres of roads and depicting street scenes at regular intervals are available (Jan, 2017). Thus OSM contributors can now use Mapillary images to map features that would require person exploration through field surveys (Juhász and Hochmair, 2016).

Even though OSM has made remarkable achievements, it still has some drawbacks. For example, according to Ibanez-Guzman et al. (2010), a high percentage of traffic accidents occur at complex and diverse road intersection areas. Therefore, as accident-prone areas, road intersections are important components of road networks, and are also quite critical to both intelligent transportation systems and navigation services.

Nevertheless, these road intersections in OSM are mainly represented as point elements without any semantic information (e.g. speed limits, turning restrictions, etc.), or are even not available for most of cities/countries (Figure 1). To the best of our knowledge, some indirect methods have been developed to detect the intersections by checking the moving direction changes of the road users on their trajectories (Karagiorgou and Pfoser, 2012; Wang et al. 2015), and a few attempts have been made to leverage high-frequency trajectory data to detect road intersections and their associated traffic rules (Xie et al. 2017). However, these methods mentioned above just detect the intersections and still represent them as a point element or an area element. As a result, the semantic information has not been complemented. Consequently, here we focus on the road intersections and propose an approach to address the problem of automated detecting and placing road objects (e.g. traffic lights and signs) using street-level images as a sole source of input data.

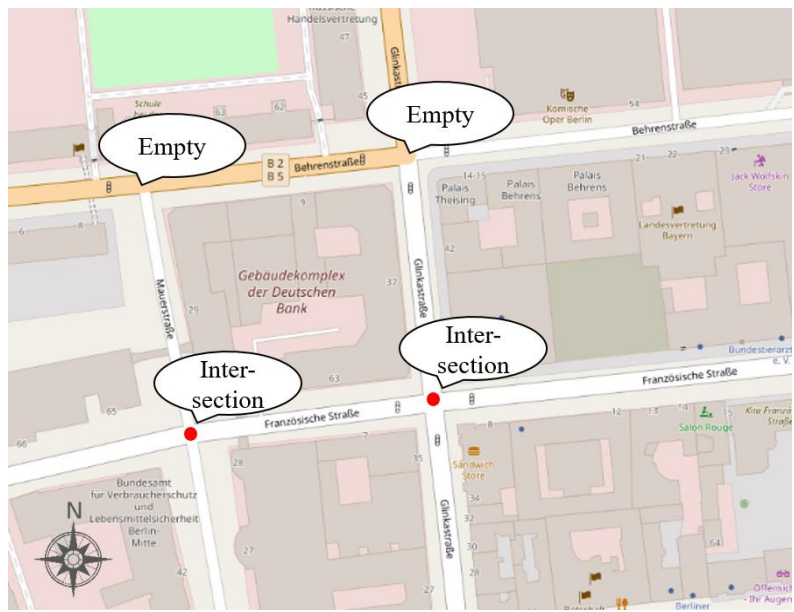

Figure 1. Incomplete information at road intersections

The remainder of this paper is organized as follows. We first review some relevant state-of-the-art approaches in Section 2. Section 3 presents our complete detection and localization pipeline. A set of experimental analyses are presented in Section 4. Conclusions and future work are discussed at the end of this paper.

**2. Related work**

Benefiting from the ubiquitous street view images accessible from Google Street View (GSV), Mapillary, etc., many efforts have been directed towards the intelligent use of them to assess urban greenery (Li et al. 2018; Li et al. 2015), to enhance existing maps with fine-grained segmentation categories (Mattyus et al. 2016), to explore urban morphologies by mapping the distribution of image locations (Crandall et al. 2009), to analyze the visual elements of an urban space in terms of human perception (Zhang et al. 2018) and urban land use (Li et al. 2017). Furthermore, street view images have also been combined with aerial imagery to achieve tree detection/classification (Wegner et al. 2016), land use classification (Workman et al. 2017), and fine-grained road segmentation (Mattyus et al. 2016). Together with (Timofte and Van Gool, 2011), these methods rely on a simplified locally flat terrain model to evaluate object locations from street-level images.

The last few years have witnessed the quick development of Convolutional Neural Network (CNN) and CNN-based image content analysis. It has been proven efficient in learning feature representations from a large-scale dataset (LeCun et al., 2015). And as a consequence, urban studies involving street-level images have been largely enhanced since it was proposed. By leveraging street view images, many studies employ deep learning for object detection and classification, as well as image semantic segmentation in order to monitor neighbourhood change (Naik et al. 2017), to quantify the urban perception at a global scale (Dubey et al. 2016), to estimate demographic makeup (Gebru et al. 2017), to predict the perceived safety responses to images (Naik et al. 2014), to predict the socio-economic indicators (Arietta et al. 2014), and to navigate without maps in a city (Mirowski et al. 2018). In contrast, less attention has been paid to extracting traffic elements within road intersections from street view imagery. What's more, all of these methods use GSV as input data, but GSV charges a fee after downloading a certain amount for free, which is no doubt not a good choice for teams or individuals with insufficient research funds. Therefore, we introduce Mapillary, a free, crowdsourced, almost real-time updated and ubiquitous street-level imagery, into our work.

In terms of localization, so far, several approaches have been made available to map particular types of objects from street view imagery: traffic lights (Jensen et al., 2016; Trehard et al., 2014), road signs (Soheilian et al., 2013), and manholes (Timofte et al., 2011). These methods determine the positions of the road assets from individual camera views based on position triangulation. All of them depend heavily on various visual and geometrical features to match when multiple objects appear in the same scene. As a result, the performance of these methods is poor when multiple identical objects exist at the same time. Thus, an improved method is proposed. Hebbalaguppe et al. (2017) describe the problem as an object recognition task, and then adopt stereo-vision (Seitz et al. 2016) approach to estimate the object coordinates from sensor plane coordinates using GSV. However, different from GSV, Mapillary street view images do not contain any camera intrinsics and projective transformation in their EXIF information, and then cannot perform the camera calibration. In other words, we cannot apply the same method for traffic lights/signs localization using Mapillary images. Recently, Krylov et al. (2018) combine use of monocular depth estimation and triangulation to enable automatic mapping of complex scenes with the simultaneous presence of multiple, visually similar objects of interest, and achieve the position precision of approximately 2 m.

In this study, we focus on the research of intersections to enrich the objects related to OSM intersections, such as traffic signs and lights, and to locate them for the reference of autonomous driving or navigation. We propose a complete pipeline to extract scene elements such as buildings, sky, roads, sidewalks, traffic lights and signs based on image semantic segmentation from road intersections images. For localization purpose, the hierarchy of semantic objects needs to be applied, as there are the coherent relations of topologies, attributes and semantics of the road objects. In further, together with the segmentation results, an attributed topological binary tree (ATBT) based on urban grammar can be established to depict the topologies among road objects. These are then matched with map features on OSM. In the end, road objects can be localized as the promising results.

## 3. Methodology

In this section, we discuss a complete pipeline for the localization of traffic lights and signs from image tracks at road intersections. The pipeline has following three modules: (1) data preprocessing and cleaning module; (2) object segmentation and recognition module; (3) localization module. Figure 2 depicts the whole framework. The first module is responsible for preparing preprocessed and cleaned data for the next two modules (see Section 3.1). The second module mainly extracts road-related information by using image semantic segmentation as well as object detection and classification (see Section 3.2). In the last module, an attributed topological binary tree (ATBT) is constructed to represent the relative position relation between extracted objects at the intersections and to locate the objects with urban grammar (see Section 3.3). Ultimately, the located objects can be integrated to enrich the OSM data.

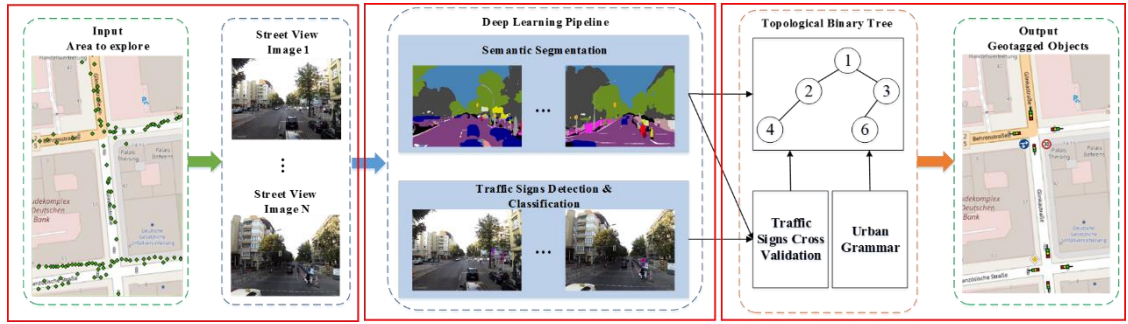

Figure 2. Workflow of the methodology

*3.1 Data Preprocessing and Cleaning*

The main purpose of data module is to prepare data for the next two modules. Specifically, all the available images can be downloaded by querying relevant Mapillary APIs, and a buffer is set up for each road intersection to extract image sequences contained in the buffer. An image sequence refers to a trajectory of a user traveling along the street. For an intersection with four road branches, we are able to theoretically build four image tracks by merging multiple image sequences according to their geolocations because of four kinds of rough driving directions, i.e. west-east, east-west, south-north and north-south. In addition, camera location, including latitude and longitude, and camera angle are extracted.

In further, we have found that the image sequences provided by Mapillary often shows the GPS position drift of the images, which may be related to the geographical environment during the shooting (for example, being around tall buildings or under heavy tree canopy that blocks the GPS signal), or it may be inaccurate with the built-in GPS receiver of the shooting device itself. Fortunately, one of the big advantages of Mapillary is that street view images of the same road segment may be uploaded repeatedly by different volunteers. And there is a certain degree of overlap between the two adjacent images, which makes it possible for us to correct their shooting positions.

Thus, in order to reduce the error as much as possible and improve the accuracy of localization, we employ a technique called Structure from Motion (SfM) (Snavely et al. 2008), as depicted in Figure 3(a), to match features between images and reconstruct their surroundings in three-dimensional space to form point clouds. Each point has its position in three-dimensional space, so we can estimate the correct shooting positions of images along with the camera angles. As a result, these corrections are capable of placing misaligned images to their original positions. In general, the more images we feed into the system from an area, the better the results can be. An SfM-corrected sequence is showed in Figure 3(b). We can easily notice the original image shooting locations (green dots) swinging from one side of the road to the other in an "S" pose. Red dots symbolize the corrected locations, which now are fully aligned with roads. Additionally, if there are many overlaps between those images, these corrections can be very promising.

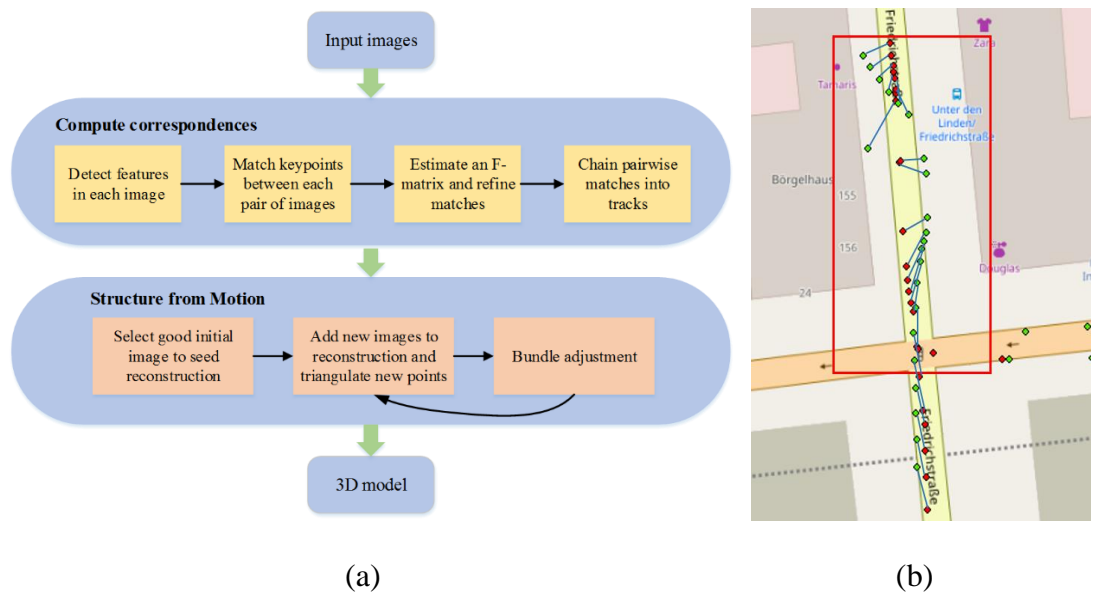

(a)                  (b)

Figure 3. Structure from Motion (SfM) algorithm used for our study to correct shooting positions of images. (a) A typical SfM pipeline (Snavely et al. 2008); (b) SfM doing its corrections

*3.2 Object Segmentation and Recognition using Deep Learning*

In theory, all road-related information can be extracted accurately from images by semantic segmentation only (see Section 3.2.1). Nevertheless, the quality of crowdsourced street-level images varies greatly, and it is difficult to ensure that all images can be segmented well, which will lead to inaccuracies or errors. Hence, in Section 3.2.2, we adopt an alternative strategy based on object detection to improve this problem.

*3.2.1 Semantic Segmentation using PSPNet*

Image semantic segmentation is one of the key techniques used to understand a scene (Zhou et al., 2017), and is aimed at segmenting and recognizing object instances from images. Given an input image, the model can assign a class label for each pixel. One of the state-of-the-art semantic segmentation models with superior performance – PSPNet (Zhao et al., 2017) is applied in our study to perform object extraction. The PSPNet uses a new neural network sub-architecture, which retains global and local contextual information through a multi-scale representation of the previous convolutional layer's output. Because of the validated performance of the PSPNet trained on the PASCAL VOC 2012 (Everingham et al., 2010) and Cityscapes (Cordts et al., 2016) datasets, we are confident to segment road-related objects well by using PSPNet, such as buildings, sky, roads, sidewalks, traffic lights and signs, etc. These extracted objects will later be used as nodes of the attributed topological binary tree (ATBT).

*3.2.2 Object Detection and Classification using YOLOv3*

Consider a street-level image where objects of interest have been segmented. We find that there are several limitations associated with semantic segmentation especially for traffic signs. Firstly, since image semantic segmentation can only detect that it is a traffic sign, but it does not know which specific kind of sign it is. Secondly, if two signs are arranged together, semantic segmentation cannot identify them separately, which is not conducive to the supplement and enrichment of OSM data. Thirdly, our PSPNet model often misclassifies the isolation piles into a traffic sign, or sometimes confuses two objects that are with similar features but not actually belong to the same class. The causes of the third limitation may be in that the features (such as color, shape or texture) of the two objects are similar, or the training dataset does not contain such cases, thus resulting in the model not learning relevant features.

Fortunately, object detection can address such limitations mentioned above. Taking into account the processing speed and detection accuracy, we choose YOLOv3 (Redmon and Farhadi, 2018) as our object detection model after some researches. Thus, we specially train a YOLOv3 model based on GTSDB (Stallkamp et al., 2012) dataset for detecting traffic signs, and then cross-validate the results of object detection and semantic segmentation to reduce errors and provide rich and effective attribute information for localization.

In our study, we not only need to know that this is a traffic sign, but also need to know which specific kind of sign it is. Consequently, in terms of traffic sign classification, we design a new shallow convolutional neural network called ShallowNet. As depicted in Figure 4, the network contains only five layers with weights; the first three are convolutional and the remaining two are fully-connected. The output of the last fully-connected layer is fed to a 45-way softmax which produces a distribution over the 45 class labels. We adopt batch normalization (Ioffe and Szegedy, 2015) right after each convolution and before ReLU non-linearity (Nair and Hinton, 2010) to speed up the convergence of model training. Additionally, in order to reduce the size of feature maps as soon as possible, the convolutional layers involved in the network are all performed filling operation, and each convolution is followed by downsampling.

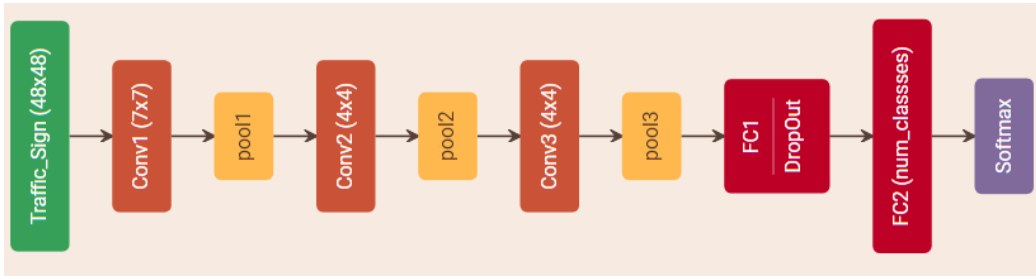

Figure 4. Overview of our proposed ShallowNet

The first convolutional layer filters the 48×48×3 input image with 64 kernels of size 7×7×3. The second convolutional layer filters the output of previous layer with 128 kernels of size 4×4×64. The third convolutional layer has 300 kernels of size 4×4×128 connected to the outputs of the second convolutional layer. Then we expand the feature map and form 1500 feature vectors into the fully-connected layer. Moreover, to reduce overfitting of the network, we introduce dropout (Hinton et al., 2012) at the first fully-connected layer.

In general, our proposed network model, ShallowNet, is characterized by:

- Simple network structure and low model complexity. With few parameters, it is easy to be deployed to mobile or embedded devices.
- High accuracy. It can correctly recognize the type of traffic signs
- Fast recognition speed. Real-time object recognition can be achieved.

*3.3 Object Localization*

Since Mapillary street view images do not contain any camera intrinsics in their EXIF information, it is impossible to calculate the projective transformation matrix and then perform camera calibration. In other words, we cannot apply photogrammetry methods for traffic lights/signs localization using Mapillary images.

After observing a large number of images of road intersections, we note that many images show a structure where buildings are on both sides of the road and a portion of sky appears between them, traffic lights and signs being often placed at street corners, as well as pedestrians and vehicles appearing on the road. We can vaguely feel that there exist some certain arrangement rules between the objects in the images. Inspired by this, we propose a novel method to depict the coherent relations of topologies, attributes and semantics of the road objects at the intersections by establishing an attributed topological binary tree based on urban grammar (see Section 3.3.1). These objects (mainly traffic lights and signs) are then further matched with map features on OSM to determine the right placed location (see Section 3.3.2).

*3.3.1 Attributed Topological Binary Tree (ATBT) Generation*

Taking the objects extracted from the image track by the semantic segmentation as input, the topological binary tree can be created from top to bottom and from left to right. The left and right children of a binary tree can reflect the relative position relationship between the objects. We regard traffic lights, traffic signs and sidewalks as three types of nodes of the tree, and assign corresponding attributes to each type of node, such as centroid, area, height (optional), category, and role in the tree.

For traffic lights, there are two types of traffic lights: located on the sidewalk (low one) and located on the road (high one), which need to be recognized through following two urban rules (see Figure 5):

(1) If the traffic light is surrounded by the sky, a ray (right red solid line in Figure 5(b)) can be casted from centroid of the segmented traffic light region downwards the road. If the distance between centroid and road surface is far more than twice height of the tallest pedestrian (blue solid line in Figure 5(b)), it can be inferred that this traffic light is at the road junction (i.e. high one), and its height is about 7 meters (*another urban rule, searched from Internet*).

(2) If the traffic light is surrounded by the buildings, a similar ray (left red solid line in Figure 5(b)) can also be casted from centroid of the segmented traffic light region downwards the sidewalk. If the distance between centroid and sidewalk surface is less than or equal to twice height of the tallest pedestrian, it can be inferred that this traffic light is located on the sidewalk (i.e. low one), and its height is about 4 meters.

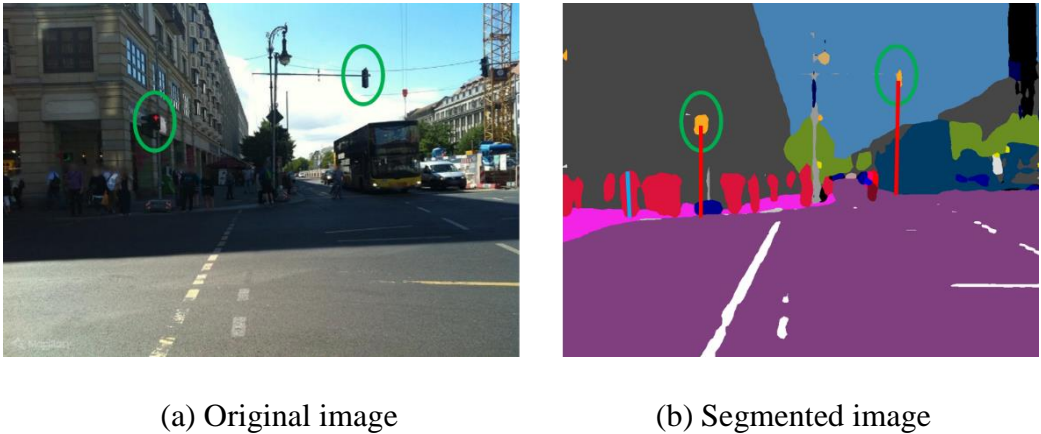

(a) Original image          (b) Segmented image

Figure 5. Discrimination of different types of traffic lights based on urban grammar, which includes two rules. **Rule1**: (1) Surrounded by sky; (2) Distance >> height of 2×max_pedestrian. **Rule2**: (1) Surrounded by buildings; (2) Distance <= height of 2×max_pedestrian.

Actually, **Rule1** implies an "up and down" relationship, that is, traffic lights are surrounded by the sky and the sky is above the high traffic lights. Similar to **Rule1**, **Rule2** also implies a "front and rear" relationship, that is, the low traffic light is surrounded by buildings and buildings are behind the low traffic light.

As we can see from **Rule2**, sidewalks are very important for our judgement. But in many cases, sidewalks are divided into multiple independent "blocks" by the pedestrian

(as shown in Figure 6(b)) according to the results of image semantic segmentation. In this case, it is necessary to judge whether the adjacent independent "sidewalk blocks" meet a certain distance threshold based on another empirical knowledge (i.e. the sidewalks on the same side are connected, *Rule3*). If within this threshold range, they are considered to be connected. What's more, many images do not capture the view of the whole intersections, but just a part of them as shown in Figure 6(a). According to the urban rules, low traffic lights on the sidewalks tend to appear in pairs (*Rule4*). As long as there is a low traffic light on one side of the road, there is definitely one on the other side of the road. This gives our topological binary tree the ability to reason.

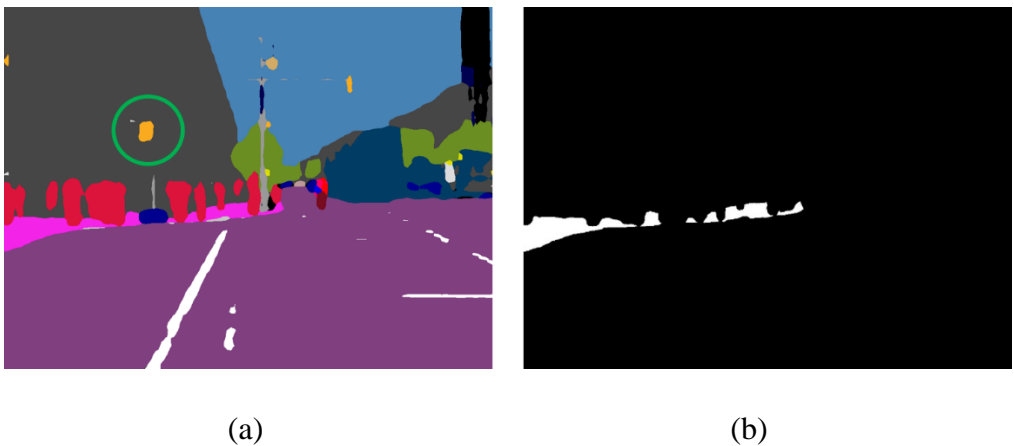

(a) (b)

Figure 6. (a) Only part of the intersection is photographed, and *Rule4* is summarized: low traffic lights appear in pairs. (b) A sidewalk is divided into multiple "blocks" by pedestrians, and *Rule3* is summarized: the sidewalks on the same side are connected.

Of course, there are also urban rules applicable to traffic signs. The area we studied is Berlin, Germany. We find that in Germany, traffic signs at intersections follow such patterns (*Rule5*, see Figure 7): they either appear alone, or are usually close to the low traffic light above or both up and down, or arrange together. These are intrinsic combination patterns, and distance between centroids of the internal objects of the combined pattern is within a small threshold.

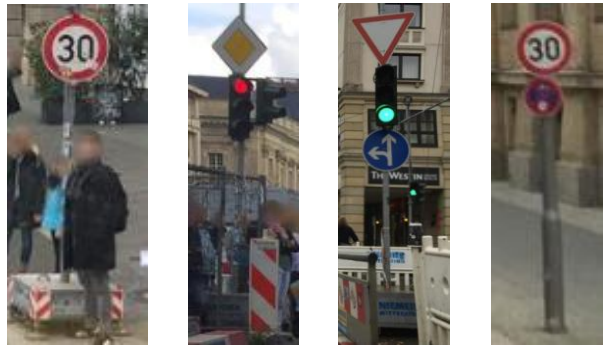

Figure 7. Four combined patterns of traffic signs and lights

Finally, for each image in the image track, an ATBT can be established from top to bottom and from left to right (as shown in Figure 8). The left subtree of root node corresponds to the left side of road, and the right subtree corresponds to the right side of road. Additionally, in order to be convenient for computation, the node number of our tree is strictly in accordance with the node number of the complete binary tree. The left-right or top-bottom relationship between nodes is determined by the position of their centroids.

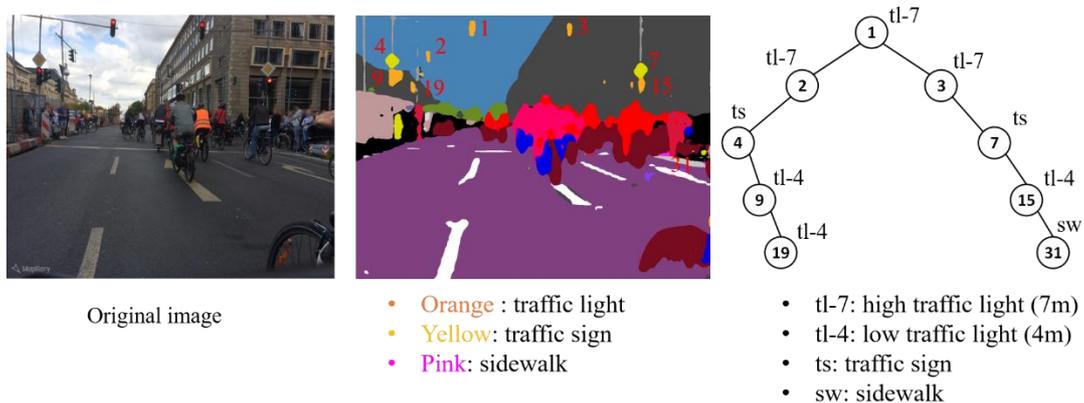

Figure 8. Attributed topological binary tree (ATBT) generation

*3.3.2 Map Matching*

Based on the ATBT constructed earlier, we can use the shooting positions and camera angles provided by images and OSM footprints at the intersections to match the left and right subtrees of the ATBT with the corresponding footprints. After that, the geographical placed locations of objects (e.g. traffic signs) in the real world can be determined.

Suppose there is an image track taken from west to east, the shooting positions of these images are no more than three typical positions represented by C1, C2 and C3 (as demonstrated in Figure 9), although this image track contains multiple images. Here, C1 is illustrated as an example. We take the red shooting point C1 as the centre of a circle, and draw the buffer with a radius of 26 meters (determined by multiple experiments) to get footprints intersecting with buffer. After calculating the distance from footprints to C1, it is found that the yellow highlighted in all right footprints is closest to the C1 (i.e. it corresponds to the right subtree of the ATBT), and similarly, the green highlighted footprint in all left footprints is closest to the C1 (i.e. it corresponds to the left subtree of the ATBT). From Figure 9, the yellow and green highlighted footprints are indeed at the intersection, which indicates that the results we got are correct. In this way, the placed positions of traffic signs and lights can be determined.

We have inquired about the "Code for Urban Road Design", which clearly states that the minimum width of an ordinary sidewalk is 2~3 meters (***Rule6***). Therefore, we place the low traffic lights and traffic signs about 2.5 meters away from the corresponding footprint corner point (A1 or A2); the high traffic lights are placed at the midpoint of connection between A1 and A2. In fact, this is not a precise localization, but it can indicate the approximate location of the traffic lights and signs.

The situation of C2 is a little bit complex. Since C2 is located at the middle of the intersection and none of footprints is around it. If C2 is the centre of a circle and the corner points obtained by intersecting with footprints are A1 and A2, it indicates that C2 just passed one side of the intersection. Because the content of an image is always the scene in front of C2, but at this time A1 and A2 are behind the C2, so these two corners are not the corner points we want. Similarly, if the circle with C2 as centre intersects with footprints and yields A3 and A4 as their corner points, which are what we want because they are in front of C2.

Compared with the situation of C2, the situation of C3 is much simpler. Since C3 is about to leave the intersection area, the corner points that C3 intersecting with footprints are always behind it. This situation is not what we want as well.

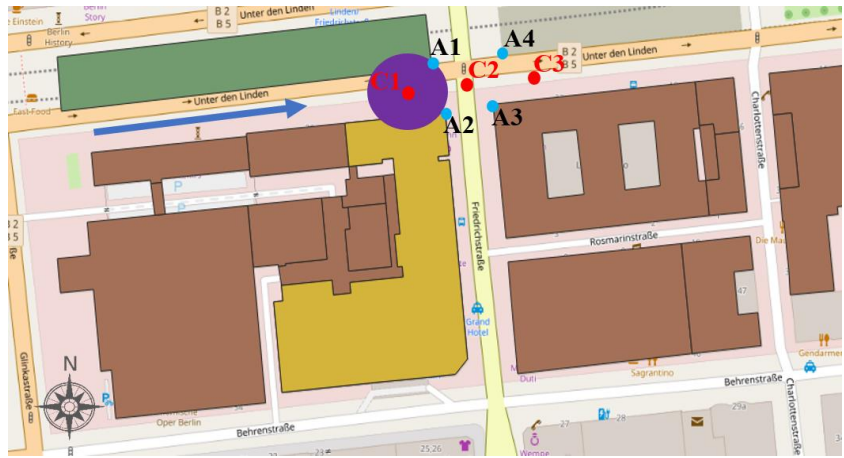

Figure 9. Map matching between shooting point and OSM footprints.

## 4. Experimental Results

### *4.1 Study Area and Data*

As the capital and largest city of Germany by both area and population, Berlin was chosen as our study area. According to recent census data (Statistical Report: Residents in the State of Berlin, 2018), Berlin has a total population about 3,740,000. The study area has varied intersection types, which range from the most common intersections with three/four road branches to the complicated intersections, like roundabouts.

The datasets used in this study include OSM building footprints data, Mapillary street view images, Mapillary Vistas, German Traffic Sign Detection Benchmark (GTSDB), and German Traffic Sign Recognition Benchmark (GTSRB) (Stallkamp et al.,

2011). The OSM building footprints data was collected from Geofabrik. The Mapillary street view images were downloaded via querying Mapillary APIs including the metadata of each image, from 2014 to 2018. In order to facilitate further study, we only extracted images located in the intersection buffer. Mapillary Vistas was from Neuhold et al. (2017), which contains 25000 high-resolution images annotated into 66 object categories. They are used as training set for the semantic segmentation model—PSPNet. Last but not the least, GTSDB and GTSRB were from Stallkamp et al. (2011, 2012), and are applied for training object detection model—YOLOv3 and proposed object classification model—ShallowNet, respectively.

In summary, above all are reasons why we choose Berlin as our study area. Figure 10 depicts the example area of Berlin as well as the distribution of Mapillary street view camera locations and OSM building footprints.

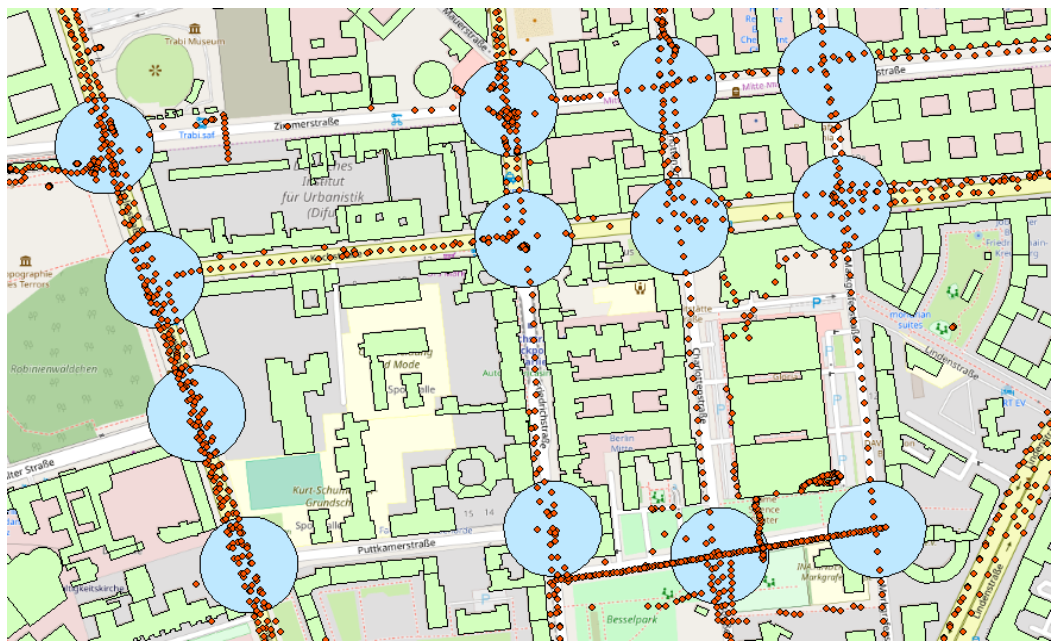

Figure 10. Example area of Berlin (map: © OpenStreetMap contributors). Red dots, green polygons, and blue circle are the Mapillary street view camera locations, OSM building footprints, and intersection buffers, respectively.

*4.2 Extraction of road-related objects with PSPNet*

**Training.** For the segmentation task, our implementation is based on the public framework TensorFlow. Like the Zhao et al. (2017), we also use the "poly" learning rate policy (the learning rate is multiplied by $\left(1 - \frac{iter}{max\_iter}\right)^{power}$) . We set base learning rate to 0.01 and power to 0.9. The training is performed on three NVIDIA GTX 1080Ti GPUs using stochastic gradient descent (SGD) with momentum m=0.99 and weight decay=0.0001. Due to limited physical memory on GPU cards, we set the "batchsize" to 4 for each GPU card during training. In addition, we crop the Mapillary training images to size of 720×720, and start with a pretrained ResNet34 (He et al., 2016) model with the dilated network strategy (Yu and Koltun, 2015) to extract the feature map. For data augmentation, we adopt random mirror, rotations [-5°, 5°], random resize between 0.5 and 2, and small enhancements in the image's color, sharpness, and brightness for Mapillary Vistas. This comprehensive data augmentation scheme makes the network resist overfitting.

**Evaluation and comparison.** The performance on Mapillary street-level images was evaluated with PSPNet. Figure 11 shows several segmented examples, where the first column represents the sample images located at the intersections, the second column corresponds to the segmented results. From the segmentation results, the PSPNet model we trained can well segment the sky, buildings, roads, traffic signs and other objects that we want. Furthermore, to prove the superiority of our PSPNet model, a comparison test is conducted with the state-of-the-art model, DeepLabv3+ (Chen et al., 2018). In Table 1, our trained PSPNet model achieves Mean IoU 34.17% and Pixel Acc. 91.3%, and both of them outperform the DeepLabv3+.

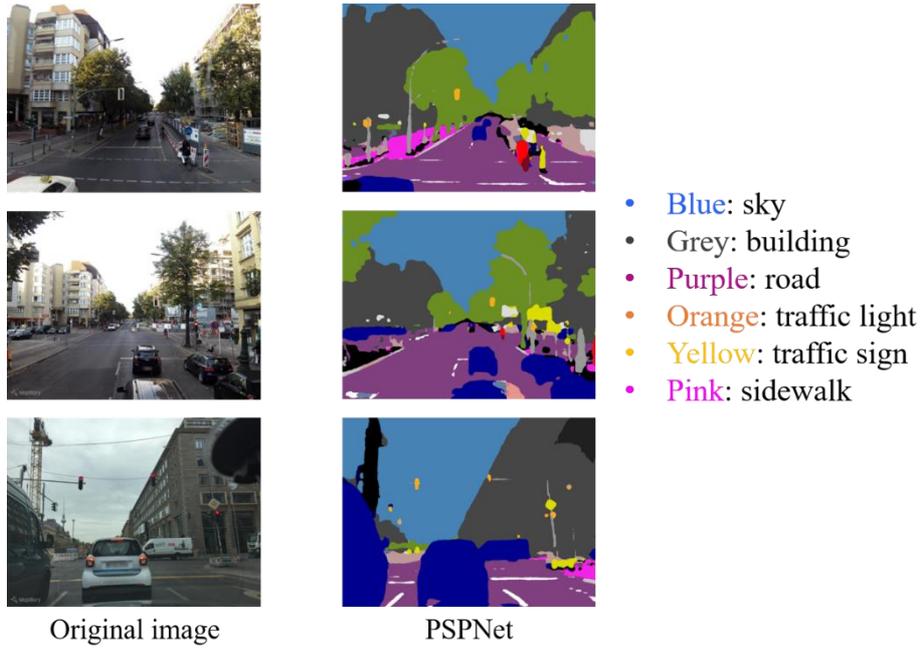

Figure 11. Examples of PSPNet results. The first column lists the original images. The second column represents the segmented results.

Table 1. Comparison of mIoU and pixel accuracy between our trained PSPNet and DeepLabv3+.

| Method | Mean IoU(%) | Pixel Acc.(%) |
|---|---|---|
| **PSPNet** | **34.17** | **91.3** |
| DeepLabv3+ | 33.97 | 90.2 |

*4.3 Detection and Classification of Traffic Signs*

In this subsection, in order to prove the superiority of our used or proposed network, we will conduct a series of comparative experiments on detection network YOLOv3 and classification network ShallowNet.

*4.3.1 Traffic Signs Detection*

**Training.** Due to some differences between the street view at intersections and the ordinary street view, we add 300 extra annotated Mapillary images at intersections into

GTSDB dataset to form a hybrid dataset. The dataset is divided into 750/450 images for training and testing. We train the YOLOv3 with Darknet on a NVIDIA GTX 1080Ti GPU card, and set "batchsize" to 8. The warm-up strategy is adopted in the training phase, i.e. starting with a very small learning rate at the beginning of training. As the number of iterations increases, the initial learning rate gradually increases to 0.001. Starting from the second epoch, the normal gradient descent is made with 0.001 as the initial learning rate. Meanwhile, to augment the data, we use rotations [-5°, 5°], random flipping, random scale [20, 200], and color space conversion.

**Evaluation and comparison.** To prove that our trained YOLOv3 model is excellent at both processing speed and detection accuracy, we compare YOLOv3 with previous best-performing method (Faster R-CNN (Ren et al., 2015)) on the testing set. In Table 2 our trained YOLOv3 model yields mAP (mean Average Precision) 94.7% and sec/img (second per image) 0.025 s, and both of them outperform the Faster R-CNN. The detection speed of approximately 30FPS is much faster than two-stage detector like Faster R-CNN. What's more, the performance of traffic sign detection on Mapillary street-level images is evaluated with YOLOv3. Figure 12 shows several example results.

Table 2. Comparison of mAP and detection time between our trained YOLOv3 and Faster R-CNN on the GTSDB + Mapillary images hybrid testing set.

| Method | Input size | mAP(%) | Model size(M) | Sec/img(s) |
|---|---|---|---|---|
| **YOLOv3(Darknet-53)** | **608×608** | **94.7** | **246.4** | **0.025** |
| Faster R-CNN(ResNet) | 1280×720 | 90.5 | 267 | 0.230 |

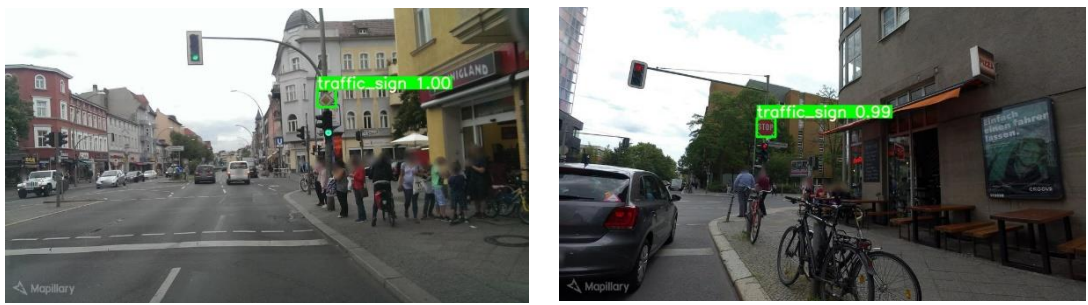

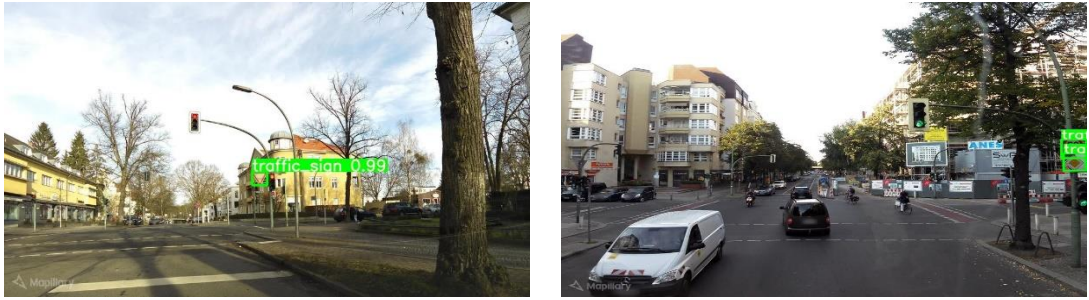

Figure 12. Examples of traffic sign detection results based on YOLOv3

*4.3.2 Traffic Signs Classification*

**Training.** The public GTSRB dataset contains only 43 types of traffic signs, but it does not cover signs that often appear at intersections. Hence, we add two more categories to reach 45 categories in total. The dataset is divided into 75K/12K images for training and testing. Due to the uneven number of different categories of traffic signs, we also use a data augmentation technique during training, which includes histogram equalization of color images, affine transformation, contrast enhancement, Gaussian blur, Gaussian random noise, color space conversion, and random inactivation of pixel values. The training is performed on a NVIDIA GTX 1080Ti GPU using Adam Optimizer with Cross Entropy Loss Function.

**Ablation Study for ShallowNet.** To evaluate ShallowNet, we conduct experiments with several settings, including batch normalization (BN), dropout, and data augmentation. As listed in Table 3, the accuracy of manual recognition is 98.84%. Although the accuracy of manual recognition is very high, the automation level is low, which is not conducive to information extraction. For the simplest ShallowNet (only convolution, pooling and full connection operation), the test accuracy on GTSRB is 95.89%. While it does not work better than manual recognition, it has higher automation level and faster forward propagation speed (it only takes 3.6ms on average to detect an image in CPU mode).

Even though the ShallowNet structure is very simple, the number of neurons in the fully-connected layer is large, which may lead to overfitting to some extent. Hence, we introduce dropout at the first fully-connected layer and successfully increase accuracy by nearly 1.6%. Besides, batch normalization is adopted in ShallowNet_Drop to reduce the difference in the distribution of original data, and to help speed up the convergence of training. ShallowNet_BN_Drop has similar performance to manual recognition. In the end, we explore whether data augmentation improves the accuracy of model or not, and augment the data on ShallowNet_BN_Drop. It achieves the accuracy of 99.52% on the testing set, which surpasses the accuracy of manual recognition, and increases by over 1% compared to ShallowNet_BN_Drop. Through this experiment, it can be proved that data augmentation is very critical to improve the accuracy of the model. Figure 13 shows several examples.

Table 3. Investigation of ShallowNet with different settings. 'Drop', 'BN' and 'Aug' represent dropout, batch normalization and data augmentation respectively.

| Method | Accuracy(%) | Sec/img(ms) |
|---|---|---|
| Human performance | 98.84 | / |
| ShallowNet | 95.89 | 3.6 |
| ShallowNet_Drop | 97.47 | / |
| ShallowNet_BN_Drop | 98.49 | / |
| **ShallowNet_BN_Drop_Aug** | **99.52** | **3.6** |

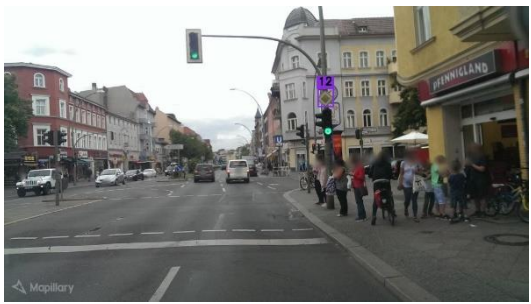
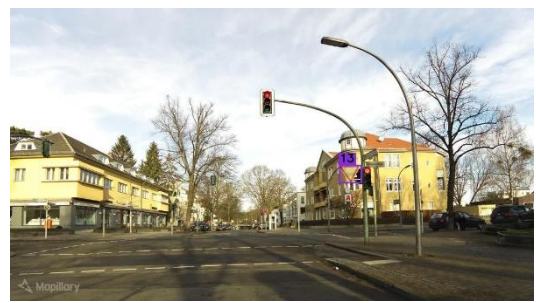

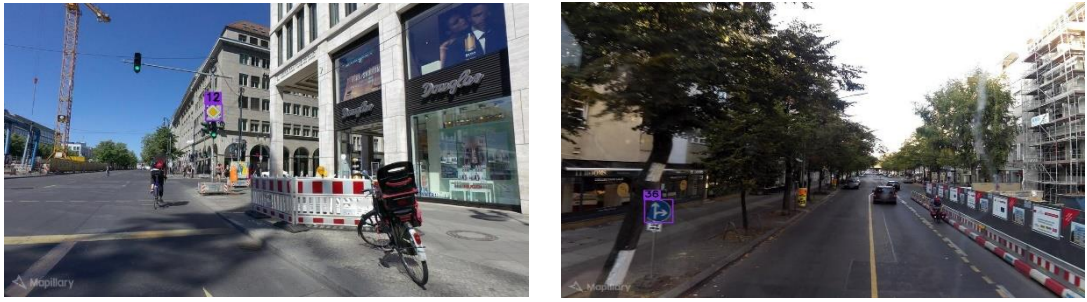

Figure 13. Examples of traffic sign classification results based on ShallowNet

*4.4 Localization of Traffic Lights and Signs*

In this section, we apply the methodology introduced in Section 3.3 for locating the traffic signs and lights using image tracks based on ATBT and urban grammar. The image track is merged from multiple image sequences according to their geolocations and meanwhile, misaligned images are corrected using Structure from Motion (SfM). Each image sequence refers to a trajectory of a volunteer user traveling along the road; and, over time, the same road segment may be covered by multiple sequences that are uploaded by different volunteers. 100 intersections (mainly crossroads and T-junctions) are tested in the experiment, with over 350 image tracks and more than 3400 images.

Using hybrid results from semantic segmentation based on PSPNet and object recognition based on YOLOv3 and ShallowNet, a parsed scene with detailed semantic and attributed information can be established. For this purpose, the hierarchy of semantic objects needs to be applied, as there are coherent relations of topologies, attributes and semantics of the road objects. Thus, one ATBT can be created based on urban grammar for each image in the image track to depict the topologies among road objects. Then, we integrate the result produced by each ATBT, rather than only using the result of one ATBT. Because it is possible that some important items (such as traffic signs) in current image are obscured by cars but the next image does not, which can play a role of verification and supplement. Ultimately, it can produce localization results along the driving direction (or camera shooting direction). Please note that this is not a precise localization, but in fact, it can indicate the approximate location of the traffic lights and signs. In Figure 14, the qualitative localization results of one crossroad and one T-junction examples are displayed.

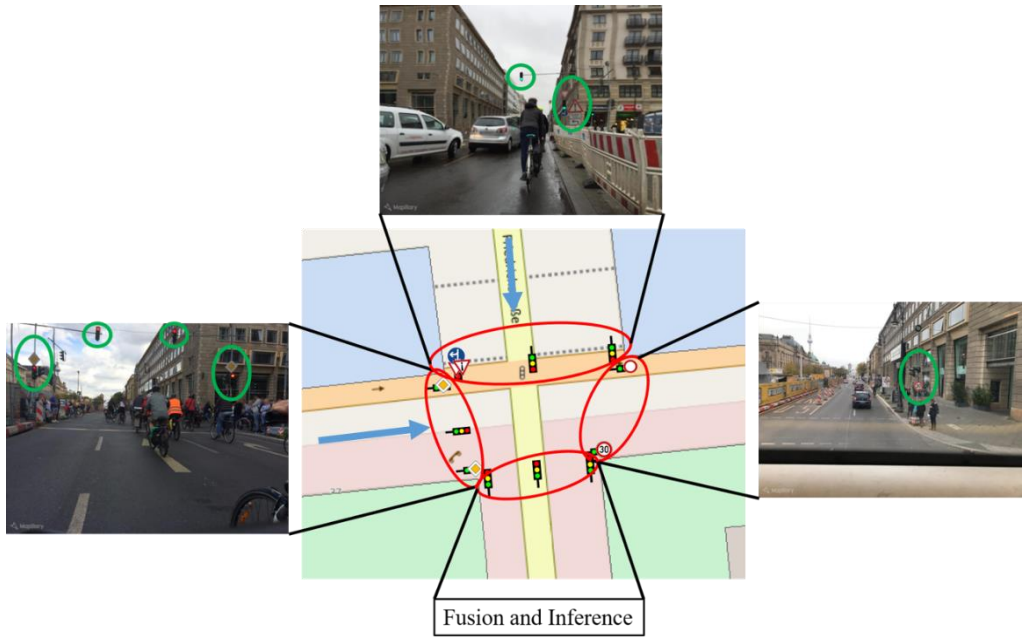

(a)

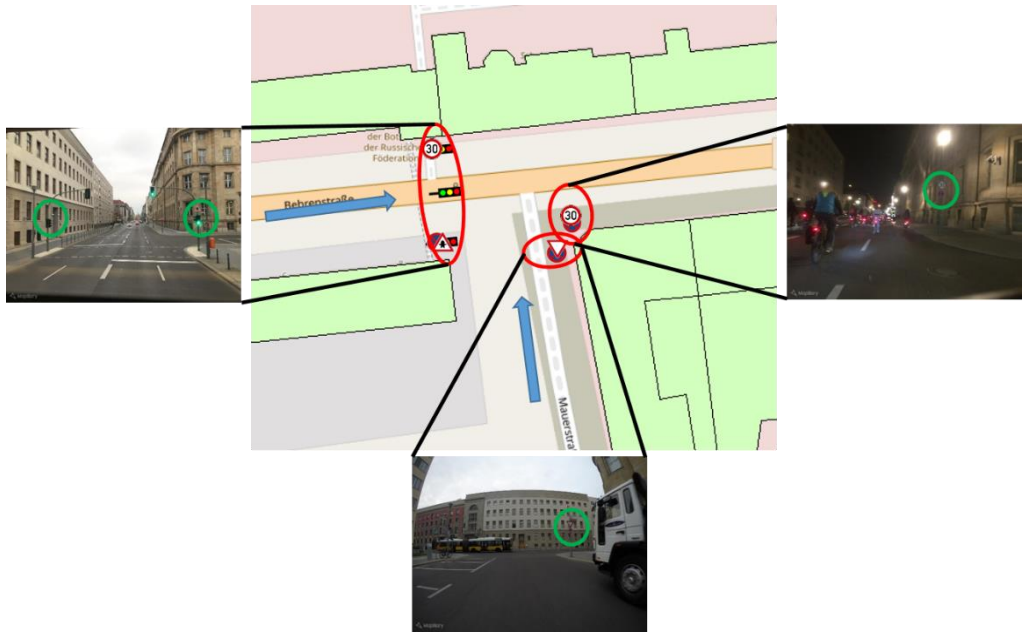

(b)

Figure 14. Examples of qualitative localization results of traffic lights and signs at two types of intersections. (a) localization results at the crossroad; (b) localization results at the T-junction

In general, for localization task, two spatial data quality elements should be assessed: completeness and positional accuracy. While positional accuracy is the best established indicator of accuracy in mapping science (Mobasheri et al., 2018), official position data (ground truth) of traffic signs and lights is not available. We cannot compare our generated positions with ground truth data on positional accuracy. However, we still manually annotate the locations of them from the street view images as another form of "reference data" to access completeness and positional accuracy of our localization results. In terms of completeness level, we get over 97% in all 100 test intersections. Figure 15 shows two examples corresponding to Figure 14a-b respectively, where red dot 1 in right figure of (a) contains three signs, and each of the red dots 1,2,3 in right figure of (b) contains two signs because they are overlapped. As can be seen in the figures, both examples have obtained approximate positional accuracy compared to the annotated "reference data".

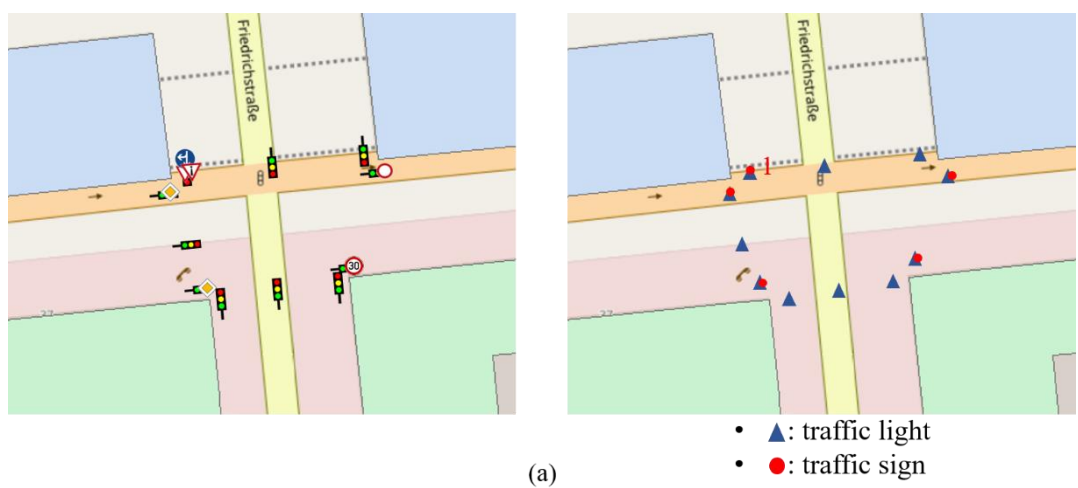

(a)

- ▲: traffic light
- ●: traffic sign

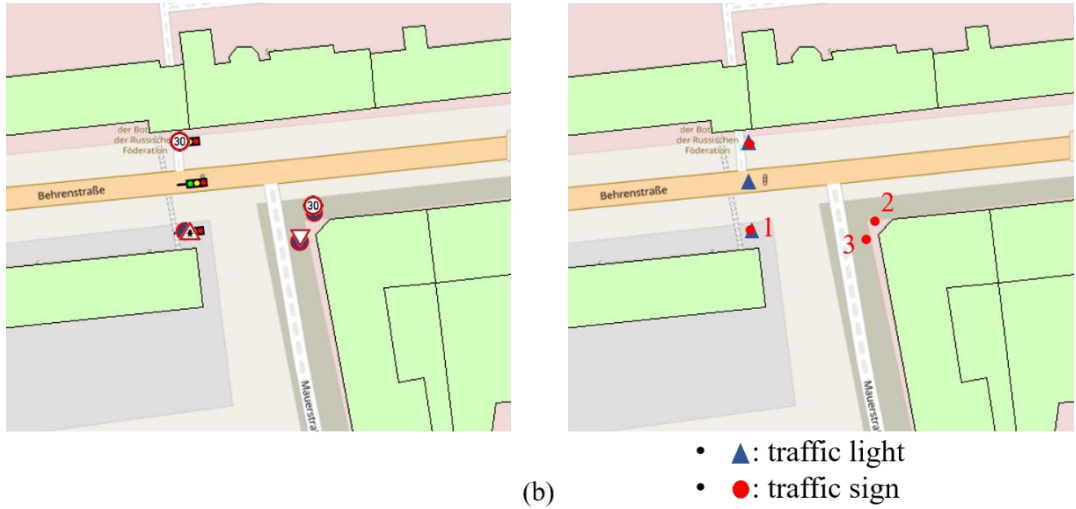

(b)

- ▲: traffic light
- ●: traffic sign

Figure 15. Visual inspection comparison of traffic lights and signs localization results. The first column is the results generated by our proposed algorithm. The second column is the annotated "reference data" we made.

## 5. Conclusions

Detailed road data are significant prerequisite for intelligent transportation systems as well as navigation. Road intersections in particular as accident-prone areas are important components of road networks, and play a critical role in route guidance. According to our investigation, road intersections in OSM are mainly represented as point elements without any semantic information (e.g. speed limits, turning restrictions, etc.), or are even not available for most cities/countries. Hence, since semantic information of intersections is not available in OSM database, to prepare data for autonomous driving or navigation service, it is required to enrich the semantic information of intersections from available sources. In this paper, we have proposed an automatic approach for detecting and placing road objects from Mapillary street-level images that are located at intersections. The proposed solution relies on two deep learning pipelines: one for image semantic segmentation and the other for object recognition. Moreover, to localize the detected objects, we establish an attributed topological binary tree (ATBT) based on urban grammar for each image in an image track. The ATBT helps to depict the coherent relations of topologies, attributes and semantics of the road objects, and simultaneously determine the right placed location of objects by matching with map features on OSM. The method has been tested at multiple intersections using Mapillary street view images in Berlin, Germany. We validate the effectiveness of our method on two object classes: traffic signs and traffic lights, and introduce two spatial data quality elements: completeness and positional accuracy. Experimental results have demonstrated that our approach obtains great objects completeness level (over 97%) and near-precise positional accuracy compared to the annotated "reference data". Therefore, the proposed method provides a promising solution for enriching and updating OSM intersection data. Among many potential applications, the output may be combined with other sources of data to guide autonomous vehicles.

At present, we only verify the effectiveness of our proposed positioning algorithm at crossroads and T-junctions. In the future, the overall robustness of the algorithm need to be enhanced by verifying at more complex intersections, like roundabouts. Since the GTSRB dataset used in this work only contains 45 categories, it does not cover all types of traffic signs in Germany or types of traffic signs in other countries. Hence, another area for future research will be the extension of GTSRB dataset to increase the generalization of ShallowNet. Ultimately, we want to create and contribute a separate intersection layer, where contains number of lanes, width of road and other road-related objects, for OSM to provide some help for autonomous driving or navigation.

Acknowledgements